\documentclass[reprint,
superscriptaddress,
amsmath,amssymb,aps,showkeys,showpacs,
twoside,final,secnumarabic,
nofootinbib]{revtex4-2}

\usepackage[paperwidth=205mm,paperheight=290mm,top=17mm,bottom=25mm,
inner=17mm,outer=17mm,
twoside]{geometry}

\usepackage{cmap} 
\usepackage[T1,T2A]{fontenc}
\usepackage[utf8]{inputenc}
\usepackage[russian,english]{babel}
\usepackage{color}
\usepackage{graphicx}
\usepackage{dcolumn}
\usepackage{bm} 
\usepackage[unicode=true,colorlinks=true,linkcolor=magenta, urlcolor=blue, citecolor = blue,breaklinks]{hyperref}
\usepackage{multirow}
\usepackage{url}
\usepackage{breakurl}
\DeclareGraphicsExtensions{.eps}

\newcount\issue
\newcount\Vol
\newcount\numb
\usepackage{fancyhdr} 
\def\Vol{\textbf{79}}
\def\numb{x}
\begin{document}

\title{Section 1. Machine learning in fundamental physics
 \\[20pt] Calibrating for the Future:\\Enhancing Calorimeter Longevity\\with Deep Learning}

\def\addressa{HSE University, Moscow, Russia}
\def\addressb{address 2}

\author{\firstname{S.}~\surname{Ali}}
\email[E-mail: ]{thraaali@hse.ru}
\affiliation{\addressa}

\author{\firstname{A.S.}~\surname{Ryzhikov}}
 \email[E-mail: ]{aryzhikov@hse.ru}
\affiliation{\addressa}

\author{\firstname{D.A.}~\surname{Derkach}}
\email[E-mail: ]{dderkach@hse.ru}
\affiliation{\addressa}

\author{\firstname{F.D.}~\surname{Ratnikov}}
\email[E-mail: ]{fratnikov@hse.ru}
\affiliation{\addressa}

\author{\firstname{V.O.}~\surname{Bocharnikov}}
\email[E-mail: ]{vbocharnikov@hse.ru}
\affiliation{\addressa}

\begin{abstract}
In the realm of high-energy physics, the longevity of calorimeters is paramount. Our research introduces a deep learning strategy to refine the calibration process of calorimeters used in particle physics experiments. We develop a Wasserstein GAN inspired methodology that adeptly calibrates the misalignment in calorimeter data due to aging or other factors. Leveraging the Wasserstein distance for loss calculation, this innovative approach requires a significantly lower number of events and resources to achieve high precision, minimizing absolute errors effectively. Our work extends the operational lifespan of calorimeters, thereby ensuring the accuracy and reliability of data in the long term, and is particularly beneficial for experiments where data integrity is crucial for scientific discovery. 
\end{abstract}

\keywords{Machine Learning, Deep Learning, Generative Adversarial Neural Networks, Calibration, High Energy Physics  \\[5pt]}

\maketitle
\thispagestyle{fancy}


\section{Introduction}\label{intro}
\subsection{Calorimeter Architecture and Energy Measurement}
A calorimeter \cite{LHCb2008, ALICE2008, ATLAS2008}is a device used in particle physics and beyond to measure the energy of incident particles. By absorbing and measuring the total energy of an incoming particle, it provides crucial data about the particle's properties \cite{Korpachov2018, Fabjan2003}. Calorimeters are essential in high-energy physics experiments, nuclear reactors, and various applications in medical physics, such as radiation therapy \cite{Korpachov2018, Fabjan2003}. They are commonly found in particle detectors like those in the Large Hadron Collider (LHC) and other major physics experiments worldwide \cite{Fabjan2003}.

Incorporating machine learning into high-energy physics (HEP) has increased for advancing our understanding of fundamental physics \cite{AI_HEP}. Machine learning (ML) techniques are transforming the way how data is analyzed and interpreted in high-energy physics experiments, especially in the wake of the Large Hadron Collider (LHC) and its upgraded version, the high luminosity LHC (HL-LHC). These experiments generate massive amounts of data, where traditional data analysis methods are insufficient for handling such large volumes, ML algorithms help in reducing data complexity and finding new features within large datasets, enhancing the overall physics performance of reconstruction and analysis algorithms \cite{whitepaper}.  ML algorithms, such as Boosted Decision Trees (BDTs) \cite{bdts} and Neural Networks (NNs), have become state-of-the-art for event and particle identification in HEP. These algorithms classify particles and events with greater efficiency and accuracy than traditional methods \cite{review}.  

Accurate simulations are essential for comparing observed data with theoretical models. ML methods, particularly deep learning models like Generative Adversarial Networks (GANs) \cite{vae_gan} and Variational AutoEncoders (VAEs), are used to generate realistic simulations of particle interactions and detector responses. These models can also accelerate computationally intensive tasks, such as event simulation and pattern recognition, which are critical for real-time data processing during experiments.

Many HEP experiments produce data at extreme rates that cannot be stored and processed offline. To manage these data volumes, real-time online data compression, zero suppression, and filtering are necessary. This real-time data processing begins at the sensor readout and extends to data storage for offline analysis. Deploying ML algorithms in custom electronics systems and edge processing on local data center servers significantly enhances the efficiency of this process, ensuring timely and accurate data handling \cite{realtime}.

ML algorithms are also essential for distinguishing these overlapping signals and isolating the rare events of interest to physicists. Techniques like pile-up suppression use ML to improve signal clarity and reduce background noise, enabling more precise measurements\cite{pileup_ML}.

\subsection{Calorimeter calibration}
Calibration ensures that the output from each cell in the calorimeter is consistent when subjected to the same input. In an ideal scenario, identical particles hitting different parts of the calorimeter should yield the same response, ensuring accurate and reliable measurements \cite{Fabjan2003}. However, due to manufacturing variations, construction misalignment and other factors each cell in a calorimeter might react slightly differently to the same input. To achieve uniformity across all cells, each one must be calibrated post-production. Calibration aligns the responses so that the measurement system can reliably report on particle interactions, which is essential for precise scientific outcomes \cite{Korpachov2018, Fabjan2003}.

\textbf{Methods of Calibration:}
\begin{enumerate}
    \item \textbf{Hardware Calibration:}
    \begin{itemize}
        \item \textbf{Internal Pulse Injection}: This involves injecting known electronic signals into the calorimeter to check its response \cite{Fabjan2003}.
        \item \textbf{Radioactive Source/Laser}: Using radioactive sources or lasers to provide a uniform and known energy input to calibrate the detector \cite{Fabjan2003}.
    \end{itemize}
    \item \textbf{Test Beam Calibration:}
    \begin{itemize}
        \item \textbf{Single Particles of a Known Type and Energy}: Single particles of a known type and energy are directed at the calorimeter to measure and adjust its response \cite{Fabjan2003}.
    \end{itemize}
    \item \textbf{In Situ Calibration:}
    \begin{itemize}
        \item \textbf{Well-Known Physics Samples}: Using particles from well-known physics processes, such as Z boson decaying to electron-positron pairs (Z→ee) or W boson decaying to jet pairs (W→jj), to calibrate the calorimeter in the actual experimental environment \cite{Fabjan2003, Korpachov2018}.
    \end{itemize}
\end{enumerate}

However, over time, the aging of components and other effects can alter the characteristics of each cell within the calorimeter, necessitating periodic recalibration to maintain accuracy and reliability. This process often requires repeating multiple calibration methods to account for changes in detector response due to radiation damage \cite{Fabjan2003}. 

Given the complexity and effort involved in optimizing a vast number of parameters during recalibration, we propose a software-based approach utilizing machine learning (ML). This method aims to automate the optimization process, reducing the need to repeat all calibration methods manually and minimizing potential issues, thereby enhancing efficiency and reliability in maintaining calibrated states of the calorimeter \cite{Fabjan2003}.

\section{\label{sec:level1}Data Description- Experimental data }
\begin{figure}
    \centering
    \includegraphics[width=1\linewidth]{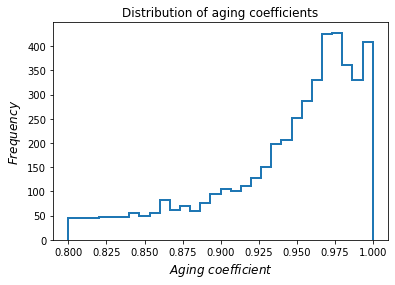}
    \caption{Distribution of aging coefficients}
    \label{fig:distOfcoeficiants}
\end{figure}

In our study, we conduct Monte Carlo simulations using the Geant4 toolkit \cite{GEANT4:2002zbu} to simulate hadronic shower samples. Specifically, we generated 10000 events with single 10 GeV pions. The simulations are carried out using a highly granular calorimeter geometry \cite{Arratia2023}, which is typical for future collider experiments. The calorimeter employed a sampling design with simplified SiPM-on-tile technology, consisting of 20mm copper plates as the absorber and polystyrene scintillator cells arranged in a 24x24 grid with a 5mm thickness as the active layers. By analyzing the energy deposition in the active cells, we aim to evaluate the efficacy of our methods by accurately capturing the changes in behavior of hadronic showers. Additionally, we present synthetic aging in the simulations to mimic radiation damage of the detector. This involved assigning an aging coefficient \( A_i \), to each readout cell \( i \), which has a linear dependence on the integrated energy deposition in each cell over the entire simulated data set as in Fig.~\ref{fig:distOfcoeficiants}

The aging coefficient \( A_i \) measures how much the response of a calorimeter cell changes over time due to factors such as radiation exposure. It reflects the difference between the cell's current state and its original, undamaged state. Specifically, for cell \( i \), let \( E_{\text{undamaged}_i} \) represent the correct energy read by the cell in its initial, undamaged state, and \( E_{\text{damaged}_i} \) denote the energy read by the same cell, from the same source of energy, after it has been damaged. The aging coefficient describes how much the current response deviates from the initial response and can be expressed as:
\[
A_i = \frac{E_{\text{undamaged}_i}}{E_{\text{damaged}_i}}
\]
If \( A_i = 1 \), it indicates no aging effects, meaning the cell's response has not changed. Conversely, if \( A_i < 1 \), it signifies a reduction in the cell's response due to damage, which necessitates calibration to ensure accurate energy measurements. By determining these coefficients for each cell, our model can make the necessary adjustments to correct the degraded responses, thereby ensuring that the calorimeter continues to provide precise energy measurements.

The dataset consists of two distinct sets: one representing energy readings from the undamaged calorimeter and the other from the damaged calorimeter. Unlike conventional machine learning applications, there is no need to split the data into separate training and test sets because our method is unsupervised. In unsupervised learning, the model is trained and tested on the same data. Our goal is to use both sets entirely to learn the aging coefficients that, when applied to one set, will match the other. Since the model does not have access to the true aging coefficients at any point and is applied in unsupervised setting, making it essential to use the full dataset to derive the best possible calibration.

\begin{figure}
    \centering
    \includegraphics[width=1\linewidth]{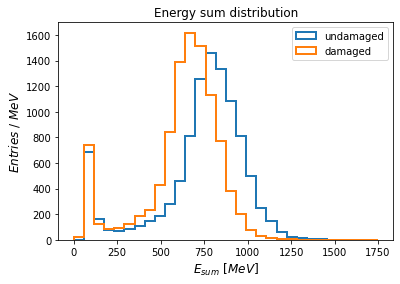}
    \caption{Energy Sum Distribution in damaged vs. undamaged Calorimeters}
    \label{fig:EnergySumDistribution}
\end{figure}

Histograms in Fig.~\ref{fig:EnergySumDistribution} of energy sum distributions for damaged and undamaged calorimeter reveal shifts in energy measurements, highlighting synthetic aging effects on calorimeter performance.

This work examines radiation damage that leads to lower signal amplitude in cells.
\section{Generative Adversarial Networks (GANs) in Calorimeter Calibration}
Generative Adversarial Networks (GANs) is a breakthrough in machine learning, particularly in the field of generative models. Introduced by Ian Goodfellow and his colleagues in 2014 \cite{GAN1}, GANs consist of two neural networks: the Generator (\(G\)) and the Discriminator (\(D\)) \cite{GAN1}. These networks engage in a minimax game, where the Generator aims to produce synthetic data that is indistinguishable from real data, and the Discriminator strives to accurately distinguish between real and synthetic data.

\textbf{Architecture and Mathematical Formulation}

The GAN framework is built upon the adversarial relationship between \(G\) and \(D\). The Generator takes a random noise vector \(z\) from a latent space and transforms it into a synthetic data sample \(G(z)\). The Discriminator, on the other hand, receives either a real data sample \(x\) or a synthetic sample \(G(z)\) and outputs a probability \(D(x)\) or \(D(G(z))\) that the input is a real sample.

The objective of the GAN training process is described by the following value function \(V(D, G)\) \cite{GAN1}:
\begin{equation}
\begin{aligned}
\min_G \max_D V(D, G) = \mathbb{E}_{x \sim p_{\text{data}}(x)}[\log D(x)] + \\
\mathbb{E}_{z \sim p_z(z)}[\log (1 - D(G(z)))]    
\end{aligned}
\end{equation}
Here, \(p_{\text{data}}(x)\) represents the distribution of real data, and \(p_z(z)\) is the distribution of the latent space from which noise vectors \(z\) are sampled. The Discriminator tries to maximize the probability of correctly identifying real and fake samples, while the Generator attempts to minimize the probability that the Discriminator accurately classifies its outputs as fake.
\subsection{Wasserstein GAN (WGAN)}
While the original GAN framework has achieved remarkable success, it often suffers from training instability and issues such as mode collapse. To address these challenges, the Wasserstein GAN (WGAN) is introduced by Arjovsky et al. in 2017 \cite{WGAN}. WGAN modifies the GAN objective by employing the Earth Mover’s distance (also known as Wasserstein distance) as a measure of the divergence between the generated and real data distributions. This change leads to a more stable training process and provides a meaningful loss metric that correlates well with the quality of generated samples.

In WGAN, the Discriminator is replaced with a Critic that provides feedback on how to adjust the synthetic data to better match the real data distribution, rather than simply classifying the data as real or fake. The WGAN objective function is expressed as:

\begin{equation}
   \min_G \max_{C \in \mathcal{C}} V(C, G) = \mathbb{E}_{x \sim p_{\text{data}}(x)}[C(x)] -\\
   \mathbb{E}_{z \sim p_z(z)}[C(G(z))] 
\end{equation}

where \(\mathcal{C}\) denotes the set of 1-Lipschitz functions. The Critic \(C\) is constrained to be 1-Lipschitz, typically enforced by clipping the weights of the Critic within a small range \([-c, c]\), gradient penalize~\cite{WGAN-GP}, or spectrally normalized~\cite{SN-GAN}. This constraint ensures that the Critic provides a reliable estimate of the Wasserstein distance between the real and generated data distributions, which helps in guiding the Generator towards producing high-quality samples.

The adoption of the Wasserstein distance in WGAN provides significant improvements in training dynamics, reducing the sensitivity to the architecture of the networks and hyperparameters, stabilizing the gradients, and often making the performance better compared to the original GAN framework. It also can be considered as a good distribution matching framework, where the WGAN Critic is responsible for the distributions (Earth Mover’s) distance assessment. 
\subsection{WGAN-inspired approach}
\textbf{Wasserstein GAN (WGAN) for Calorimeter Calibration}

\begin{figure}
    \centering
    \includegraphics[width=1\linewidth]{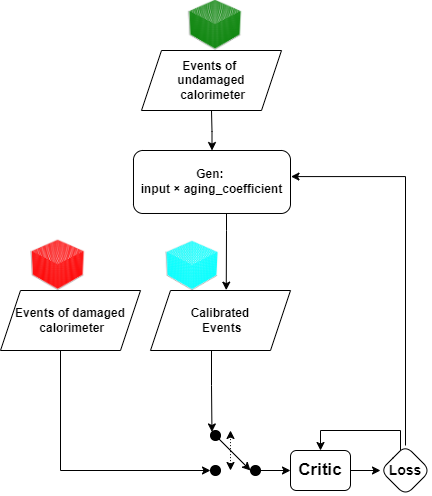}
    \caption{Architecture of the WGAN-inspired network used for calibrating the calorimeter.}
    \label{fig:archMod}
\end{figure}

WGAN works by minimizing the Wasserstein distance (Earth Mover’s distance) between the generated data distribution and the real data distribution. This is achieved through having the loss for the Critic as the Wasserstein distance. Inspired by this approach, we employed WGAN architecture to calibrate a calorimeter response, aiming to align the distribution of data from a undamaged calorimeter with that from damaged calorimeter.
In our application, the Generator's weights represent the predicted aging coefficients, and the synthetic data processed by the Generator corresponds to the corrected data from the undamaged calorimeter. 
The Critic is trained to estimate the Wasserstein distance, while the Generator iteratively updates its weights to reduce this distance, learning the appropriate aging coefficients. This process results in a corrected data distribution that closely matches the damaged calorimeter data. Unlike traditional WGANs focused on generating new data, our WGAN-inspired approach emphasizes aligning existing data distributions. To ensure the model has the best potential to learn the aging effects, we conducted the training specifically on the most affected cells located in the center of the calorimeter in our dataset, while not focusing on the cells at the borders of the calorimeter. The architecture of our WGAN network, including the Generator and Critic components, is illustrated in Fig~\ref{fig:archMod}.

\begin{figure}
    \centering
    \includegraphics[width=1\linewidth]{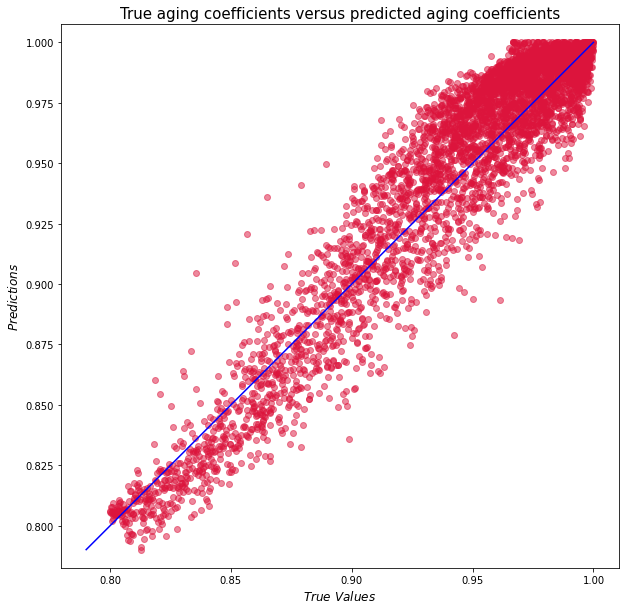}
    \caption{Scatter plot of true aging coefficients versus predicted aging coefficients. The closer to line, the more consistent with true values the predictions are.}
    \label{fig:predicted_vs_true}
\end{figure}

\begin{figure}
    \centering
    \includegraphics[width=1\linewidth]{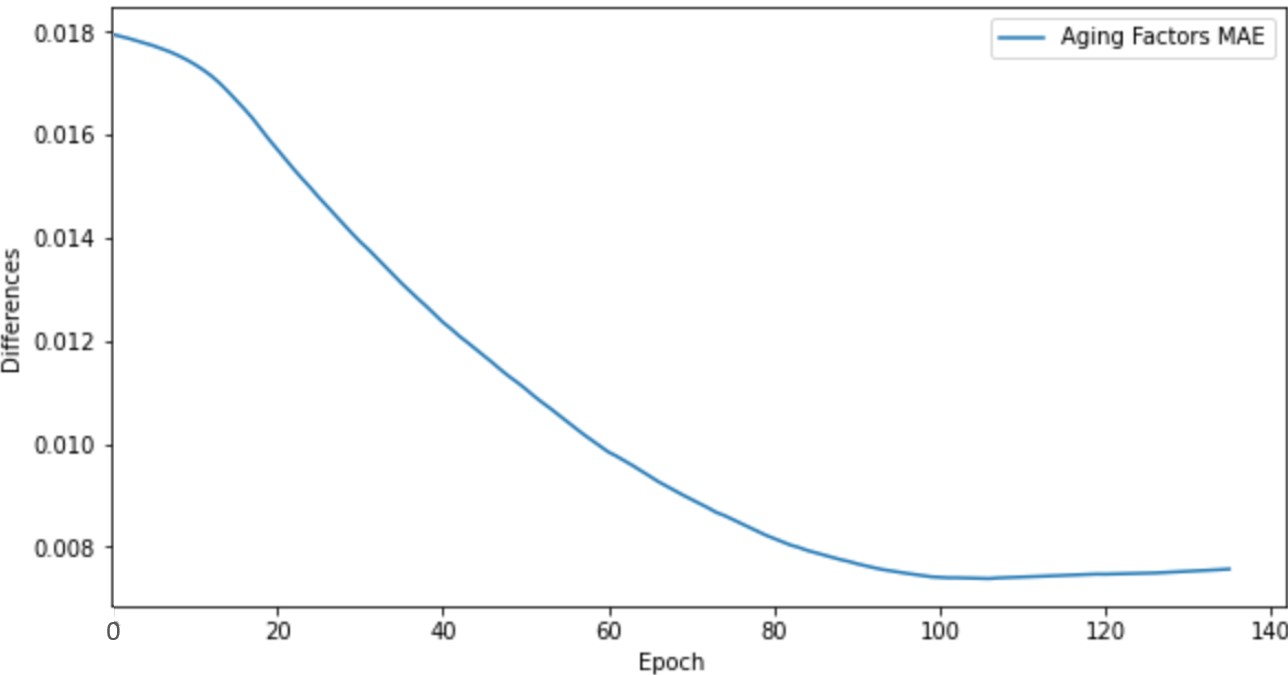}
    \caption{Mean Absolute Error (MAE) of aging coefficients. The MAE decreases consistently, stabilizing around 0.0074 after 100 epochs, with R\textsuperscript{2} value of 0.88, demonstrating the model's potential to tune aging coefficients.}
    \label{fig:mae_during_training}
\end{figure}

\begin{figure}
    \centering
    \includegraphics[width=1\linewidth]{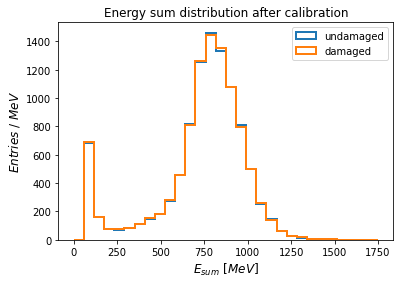}
    \caption{Energy sum distribution for the undamaged and damaged calorimeters. As Fig \ref{fig:EnergySumDistribution} shows a visible shift between the two distributions before calibration while here we see the distributions after calibrating the damaged calorimeter with our predicted aging coefficients, indicating a better alignment. Calibration is done at the cell level.}
    \label{fig:energy_sum_distribution_after_calb}
\end{figure}

\section{Results and Discussion}
The results of the predicted aging coefficients by our WGAN-inspired approach are illustrated in Fig. \ref{fig:predicted_vs_true}. The scatter plot of the true values versus the predicted values shows that the points are closely aligned along the diagonal line. This indicates that the model has a strong capability to predict the aging factors accurately. Moreover, Fig. \ref{fig:mae_during_training} demonstrates the Mean Absolute Error (MAE) of the aging factors throughout the training process. The MAE decreases consistently, starting from 0.018 and stabilizing around 0.0074 after 100 epochs. This consistent reduction in error showcases the model's effectiveness in tuning the aging coefficients.

Additionally, the alignment of the scatter plot points along the diagonal line suggests that the model captures the underlying distribution of the data. The continuous decrease in MAE further reinforces the reliability of the model's predictions over time. These results demonstrate the potential of the inspired-WGAN approach to effectively calibrate the calorimeter by adjusting the aging coefficients to closely match the distribution of data from the undamaged calorimeter. The improvement in MAE over epochs highlights the model's learning capability and the ability to reduce prediction errors, thus providing potential in its application for real-world scenarios.

Fig. \ref{fig:energy_sum_distribution_after_calb}  illustrates the performance of our model using the predicted aging coefficients to calibrate the same dataset used for training. Initially, there is a visible shift between the two distributions, indicating significant differences due to the damage. After applying our predicted aging coefficients and calibrating the damaged calorimeter, the distribution of the calibrated data aligns much more closely with that of the undamaged calorimeter. which also demonstrates the potential of our calibration method. Calibration is performed at the cell level, and the damaged calorimeter shows fewer changes in its distribution post-calibration, indicating a successful adjustment of the aging coefficients.
\section{Conclusion}
Generative machine learning models have demonstrated their ability to address degradation effects in particle detectors, providing an innovative approach to calibration challenges. Our WGAN-inspired model has shown significant potential in identifying calibration coefficients. In our study, the model achieved a Mean Absolute Error (MAE) of 0.0074, using a dataset of 5000 events for each damaged and undamaged states.

The efficiency of our WGAN model translates into substantial benefits, potentially reducing the cost and resource intensity of calibration experiments. By enabling calibration with a reduced amount of experimental data, this approach offers a more practical and economical solution for maintaining the accuracy of particle detectors over time. 

Moreover, the presented method is universal and adaptable, capable of being applied to detectors of various shapes and based on different physics principles. This versatility underscores the broad applicability and impact of our approach, paving the way for its use in a wide range of detector calibration scenarios. Our work highlights the promise of generative models in enhancing the longevity and performance of particle detectors, contributing to the advancement of experimental physics.
\section*{FUNDING}
This work is supported by HSE Basic research fund. The computation for this research was performed using the computational resources of HPC facilities at HSE University.



\end{document}